\documentclass{article}

\usepackage{microtype}
\usepackage{graphicx}
\usepackage{subcaption}
\usepackage{booktabs} 

\usepackage{hyperref}

\usepackage[accepted]{icml2026}

\usepackage{amsmath}
\usepackage{amssymb}
\usepackage{mathtools}
\usepackage{amsthm}
\usepackage{listings}
\usepackage{xcolor}

\theoremstyle{plain}

\theoremstyle{definition}

\theoremstyle{remark}

\lstdefinelanguage{Dafny}{
  morekeywords={array,forall,if,method,return,returns,var,while},
  morekeywords=[2]{ensures,requires},
  morekeywords=[3]{assert,invariant},
  sensitive=true,
  morecomment=[l]{//},
  morestring=[b]"
}
\icmltitlerunning{AxDafny}

\begin{document}

\twocolumn[
  \icmltitle{AxDafny: Agentic Verified Code Generation in Dafny}

  \begin{icmlauthorlist}
    \icmlauthor{Benjamin Breen}{axiomatic}
    \icmlauthor{Austin Letson}{axiomatic}
    \icmlauthor{Borja Requena Pozo}{axiomatic}
    \icmlauthor{Leopoldo Sarra}{axiomatic}
  \end{icmlauthorlist}

  \icmlaffiliation{axiomatic}{Axiomatic AI, Boston, MA, USA}

  \icmlcorrespondingauthor{Benjamin Breen}{ben@axiomatic-ai.com}

  \icmlkeywords{Formal Verification, Dafny, Agentic Systems, Code Generation, Benchmarks}

  \vskip 0.3in
]



\printAffiliationsAndNotice{}  

\begin{abstract}
We study agentic code generation in Dafny, where a model must generate both executable code and the proof artifacts for verification. We present AxDafny, a verifier-guided repair framework that iteratively generates implementations, invariants, assertions, and termination arguments. We also
introduce LiveCodeBench-Pro-Dafny (LCB-Pro-Dafny), a benchmark of 250
competition-style programming problems translated into Dafny with formal specifications and a verifier-based evaluation harness. On LCB-Pro-Dafny, AxDafny substantially improves verification success over baseline GPT-5.5 performance. On DafnyBench, AxDafny verifies 725/782 instances (92.7\%), outperforming the strongest previously reported proof-hint baseline by 6.5 percentage points. Lastly, we show that verification success and runtime test performance measure different aspects of generated code.
\end{abstract}

\section{Introduction}

Inference scaling for coding agents often takes the form of an iterative
generate--check--repair loop: a model proposes code, receives feedback from an environment, and then refines the code based on the feedback 
\citep{yang2024sweagent,requena2026minimalagentautomatedtheorem}.
Formal methods provide a rich feedback signal: rather than checking
a fixed set of test cases, a verifier can identify when a program 
violates its specification, when additional proof annotations are needed to verify logical steps, 
and when edge cases are not handled. This makes verified code generation a practical setting for studying the effects of inference scaling in agentic coding. Despite this promise, applications of formal methods in mainstream AI tasks remain
limited due to the substantial expertise they require.

We study how verification feedback can improve agentic code generation, using Dafny as an experimental testbed because it provides built-in support for formal verification. We make two main contributions. First, we introduce AxDafny, an agentic framework for program and proof synthesis that generates verified implementations through iterative verifier-guided repair. Second, we introduce LiveCodeBench-Pro-Dafny (LCB-Pro-Dafny), a benchmark that translates competition-style coding problems into Dafny to evaluate both program synthesis and proof synthesis. 

Empirically, AxDafny with Gemini-3.1-Pro verifies 92.7\%
on the established DafnyBench proof-hint task, outperforming the strongest previously reported baseline by 6.5 percentage points. On the newly introduced LCB-Pro-Dafny benchmark, AxDafny verifies 56.4\% of tasks, improving over the 11.6\%
success rate of direct GPT-5.5 generation. A secondary evaluation compiles
verified LCB-Pro-Dafny solutions to Python and executes them under the
original LiveCodeBench-Pro test harness. We observe that most executable
failures result from resource limits, because the Dafny specifications
enforce functional correctness rather than asymptotic complexity. Together, these results demonstrate the efficacy of verifier-guided repair for functional
verification, while exposing the remaining challenge of generating verified
code that meets competitive-programming resource constraints.

\subsection{Dafny and Formal Verification}

Dafny is a programming language with built-in support for formal verification
\citep{leino2010dafny}. A Dafny method can
state assumptions about its inputs using \texttt{requires} and guarantees about its outputs using \texttt{ensures}; these are known as
preconditions and postconditions, respectively. Dafny checks that the implementations satisfy
these specifications using an SMT-based verifier, reporting failed proof obligations with source locations and diagnostic messages.

Verification in Dafny often requires additional annotations beyond the executable
code (Figure~\ref{fig:dafny-annotation-example}). Loop invariants specify conditions that remain true across every iteration of a \texttt{for} or \texttt{while} loop; for example, an invariant in a sorting loop might state that the processed prefix is sorted. Assertions ask Dafny to prove a local fact at a specific point in a program; for example, \texttt{assert a[j] <= a[j+1]} can expose an ordering fact needed to preserve the loop invariant or
prove the final postcondition. Finally, termination
arguments show that recursive calls or loops make progress. 

Dafny provides two additional properties useful for code generation. First, it
can produce concrete counterexamples for certain failed verification conditions.
Second, verified programs can be compiled to standard programming languages
such as Python, Java, and C++ for execution.

These properties make Dafny a strong setting to study verifier-guided LLM code generation: verifier failures  identify both program errors and missing proof
obligations, while successful outputs remain executable. Unlike ordinary code generation, Dafny generation
requires the model to produce both the implementation and the proof structure
needed for verification. We therefore distinguish \emph{program synthesis},
which generates executable Dafny code, from \emph{proof synthesis}, which
generates invariants, assertions, and other annotations needed to establish
verification.

\begin{figure}[t]
\begin{lstlisting}[language=Dafny]
method Sum(n: int) returns (s: int)
  requires n >= 0
  ensures s == n * (n + 1) / 2
{
  var i := 0;
  s := 0;

  while i < n
    invariant 0 <= i <= n
    invariant s == i * (i + 1) / 2
  {
    i := i + 1;
    s := s + i;
  }

  assert i == n;
}
\end{lstlisting}
\caption{A verified Dafny method for computing an arithmetic sum from 0 to $n$.
Executable statements, highlighted in blue, implement the main loop.
Method specifications, highlighted in green, use \texttt{requires} and
\texttt{ensures} to specify input assumptions and output guarantees. Proof
annotations, highlighted in orange, use \texttt{invariant} and \texttt{assert}
to supply intermediate facts to the verifier in order to prove the postcondition.}
\label{fig:dafny-annotation-example}
\end{figure}

\section{Related Work} 

\subsection{Agentic Architectures for Dafny}

Several recent Dafny systems study LLM-assisted generation with verifier
feedback. DafnyBench evaluates LLM proof-hint generation with a basic
iterative loop: the model proposes annotations, Dafny checks the file, and
verifier errors are returned for repair \citep{loughridge2024dafnybench}.
DafnyPro improves this setting by editing
existing files instead of regenerating whole proofs, pruning unused invariants, and retrieving from a manually
curated Dafny proof-hint library \citep{banerjee2026dafnypro}. 

These systems target proof generation, whereas our work addresses the joint synthesis of programs and proofs. Moreover, our method does not rely on proof-hint retrieval to improve proof-synthesis performance. 

\subsection{Dafny Benchmarks}

Most public Dafny benchmarks target proof synthesis rather than program
synthesis. DafnyBench removes \texttt{assert} and \texttt{invariant}
annotations from existing verified programs and evaluates whether a model can
restore sufficient proof hints for verification \citep{loughridge2024dafnybench}.
DafnyComp evaluates generation of specifications and proof annotations across
multiple Dafny programs \citep{xu2025dafnycomp}. VeriEquivBench evaluates two
related proof tasks: refining proof annotations for a fixed specification, and
generating the strongest specification for a fixed implementation
\citep{zeng2026veriequivbench}.

Public benchmarks for Dafny program synthesis are less common. Dafny-synthesis
evaluates verified Dafny method generation from natural-language prompts
\citep{misu2024towards}; its 178 tasks are derived from the Mostly Basic
Python Problems (MBPP) benchmark \citep{austin2021program}. HumanEval-Dafny
translates 129 tasks from the original HumanEval benchmark into Dafny
\citep{chen2021evaluating,jetbrains2024humanevaldafny}. These benchmarks are
limited for comparisons with mainstream programming languages by their scale, simplicity, potential
data contamination, and lack of runtime constraints. We
therefore introduce LCB-Pro-Dafny, a larger competition-style benchmark for
verifier-guided Dafny program synthesis.

\section{Methods}

\subsection{Agent Architecture}

For AxDafny, we adapt the agentic architecture proposed in \citet{requena2026minimalagentautomatedtheorem} used for Lean proving to Dafny code synthesis and verification. It consists of an iterative refinement loop with a proposer agent, a review system, and a memory module. 
The proposer writes Dafny code, the review system checks it and returns concrete feedback, and the memory module preserves information across iterations.

\paragraph{Proposer.} The proposer's goal is to produce a Dafny file that passes \texttt{dafny verify}, without modifying the code and \texttt{ensures} statements that define the problem. 
It receives the original file content, the name of the primary target, and any context from previous attempts maintained by the memory module. 
With this information it returns the complete annotated file.

\paragraph{Reviewer system.} The reviewer has two stages: deterministic checks and LLM-based review. The deterministic stage consists of two static checks on the proposed file, followed by verifier diagnostics from Dafny.

First, we extract the \texttt{requires} and \texttt{ensures} clauses from the original and proposed files, and require the original clauses to be a subset of the proposed clauses. This prevents specification weakening while still allowing new specifications for helper methods and lemmas.

Second, a regex filter rejects proof-bypass constructs: \texttt{\{:axiom\}} and \texttt{\{:verify false\}} for specifications, \texttt{assume} for asserting verification facts without proof, and \texttt{\{:extern\}} for declarations whose implementations are assumed to be verified externally. If either check fails, the offending lines are returned to the proposer.

We then run \texttt{dafny verify} on the generated file. When verification fails, diagnostics identify the failed proof obligations and their source locations, including unmet preconditions, postconditions, invariants, or termination checks. When enabled, the \texttt{--counterexamples} flag can also provide concrete counterexamples for invalid verification conditions. These verification messages are fed back to the proposer as repair feedback.

Finally, a reviewer LLM checks the file for cheating patterns which aren't covered by the deterministic checks, for example rewriting a predicate to be trivially true. A candidate is accepted only if it passes the preservation checks, contains no
rejected proof-bypass constructs, verifies with Dafny, and is not rejected by the LLM reviewer.

\paragraph{Context management.} We use a self-managed, reflection memory variant as in \citep{requena2026minimalagentautomatedtheorem}, adapted to Dafny verifier output. On every iteration the proposer receives two forms of historical context. First, the immediately previous attempt is included verbatim: the proposer's prior reasoning, the proposed file, and the feedback returned by either \texttt{dafny verify} or the reviewer agent. Second, a separate scratchpad call maintains a rolling memory that condenses lessons accumulated across all earlier attempts. After each iteration, this memory is regenerated by an LLM given the current attempt's reasoning, code, and feedback together with the previous notebook contents, with explicit instructions to preserve earlier lessons while folding in new ones. 

\subsection{Agent Configuration}

Experiments use either Gemini-3.1-Pro or GPT-5.5 as the base language model. GPT-5.5 is evaluated with either the low or medium reasoning setting. In small-scale ablations, Gemini-3.1-Pro and GPT-5.5 provided the best performance-cost tradeoffs among the models tested. For the full evaluation, AxDafny was allowed a budget of 20 iterative attempts per benchmark instance.

\subsection{Benchmarks}

We evaluate AxDafny on two complementary verified-code tasks. DafnyBench
measures proof-hint synthesis: the implementation and specification are
given, and the agent must restore missing invariants and assertions so
that Dafny verifies the file \citep{loughridge2024dafnybench}. LCB-Pro-Dafny, introduced in this work, measures program synthesis: the agent receives both a natural-language problem description and Dafny specification, and must generate the implementation and proof artifacts needed for verification.

\paragraph{LCB-Pro-Dafny Construction}

LCB-Pro-Dafny contains 250 LiveCodeBench-Pro
\citep{zheng2025livecodebenchpro} competition problems translated into Dafny:
100 easy, 100 medium, and 50 hard instances. Each instance contains the original
natural-language problem statement, a Dafny method signature, and a formal
specification. The benchmark is released with an evaluation harness; acceptance
criteria are defined in Section~\ref{sec:metrics}.

\paragraph{Benchmark validation.}
Specification quality is a known bottleneck for verifiable-code benchmarks
\citep{zeng2026veriequivbench}. We therefore treat LCB-Pro-Dafny construction
as a curation process rather than a fully automatic translation step. Each
specification was checked by LLM-as-judge review against the original problem
statement, verifier-based sanity checks, and manual review of flagged cases. All
250 specifications parse and type-resolve under Dafny 4.11.0. LLM-as-judge
review flagged 14/250 specifications (5.6\%) for semantic fixes; all flagged
specifications were revised before inclusion. Appendix~\ref{app:lcb-pro-dafny-construction}
reports the validation checks and counts.

\subsection{Metrics}
\label{sec:metrics}

For proof synthesis on DafnyBench, we follow the benchmark's original
evaluation protocol \citep{loughridge2024dafnybench}. A solution is
accepted only when it verifies in Dafny and does not alter the
\texttt{requires} or \texttt{ensures} clauses or rely on vacuous proofs such
as \texttt{assume false} or \texttt{\{:verify false\}}.

For program synthesis on LCB-Pro-Dafny, we use an analogous evaluation
protocol. A solution is accepted only when (1) the generated implementation verifies in Dafny, (2) the provided specification, definitions, and functions are not modified, and (3) the solution does not rely on vacuous proof shortcuts, including \texttt{assume} or \texttt{\{:extern\}}.

Additionally, we report a secondary functional-correctness metric using
the original LCB-Pro test cases \citep{zheng2025livecodebenchpro}. For
this evaluation, the generated Dafny code is compiled to Python and
executed against the corresponding test suite. This metric is separate from
verifier acceptance and measures executable correctness under the original
competitive-programming harness.

\subsection{Baselines}

For DafnyBench, we compare with pass@1 model results and published proof-hint results using the original evaluation protocol, including DafnyBench, dafny-annotator, and DafnyPro \citep{loughridge2024dafnybench,
poesia2024dafnyannotator,banerjee2026dafnypro}. These baselines provide
context for the current frontier on the established proof synthesis task.

For LCB-Pro-Dafny, our primary baseline is GPT-5.5 pass@1. The model is
given the same problem statement and Dafny specification as AxDafny, but
is evaluated using a single non-iterative generation attempt. This baseline measures the verification success of direct model output without agentic verifier-guided repair.

As context for cross-language baselines, Appendix~\ref{app:lcb-pro-standard-results}
reports standard LCB-Pro leaderboard results under the original Python
test-based evaluation. These results are not directly comparable to
LCB-Pro-Dafny verification rates, but they contextualize the difficulty of the
programming tasks under the original runtime-test criterion.

\section{Results}

We report verification success on DafnyBench and LCB-Pro-Dafny. The
LCB-Pro-Dafny results are stratified by difficulty to separate performance on
easy, medium, and hard instances.

On DafnyBench, Table~\ref{tab:dafnybench-results} shows that AxDafny with Gemini-3.1-Pro verifies 725\,/\,782 benchmark instances (92.7\%), improving over the strongest previously reported proof-hint baseline by 6.5 percentage points. The comparison with DafnyPro requires some additional context: DafnyPro reports 86.2\% for its full system, which includes a retrieval component from a manually curated
proof-hint library. Its ablation attributes 9.97 percentage points of
DafnyBench performance to hint augmentation; without this component, DafnyPro
reports 76.2\% success. AxDafny does not use a task-specific proof-hint library.

While Table~\ref{tab:dafnybench-results} reports the standard evaluation setting for the DafnyBench proof-hint task, we also test a more permissive setting in which models may introduce
helper lemmas and propositions in addition to proof annotations. Under this
evaluation setting, AxDafny with GPT-5.5 medium verifies 92.0\% of DafnyBench instances.

\begin{table}[!htbp]
  \caption{DafnyBench proof-hint results.}
  \label{tab:dafnybench-results}
  \centering
  \begin{tabular}{lc}
    \toprule
    Method & Verified (\%) \\
    \midrule
    \textbf{AxDafny (Gemini-3.1-Pro)} & \textbf{92.7} \\
    \textbf{AxDafny (GPT-5.5 medium)} & \textbf{88.9} \\
    \midrule
    DafnyPro \citep{banerjee2026dafnypro} & 86.2 \\
    \textbf{AxDafny (GPT-5.5 low)} & \textbf{85.3} \\
    DafnyPro w/o hints & 76.2 \\
    Gemini-3.1-Pro pass@1 & 68.5 \\
    DafnyBench \citep{loughridge2024dafnybench} & 68.0 \\
    dafny-annotator \citep{poesia2024dafnyannotator} & 50.6 \\
    GPT-5.5 (medium) pass@1 & 54.6 \\
    GPT-5.5 (low) pass@1 & 54.1 \\
    \bottomrule
  \end{tabular}
\end{table}

\begin{figure}[!t]
  \centering
  \includegraphics[width=\columnwidth]{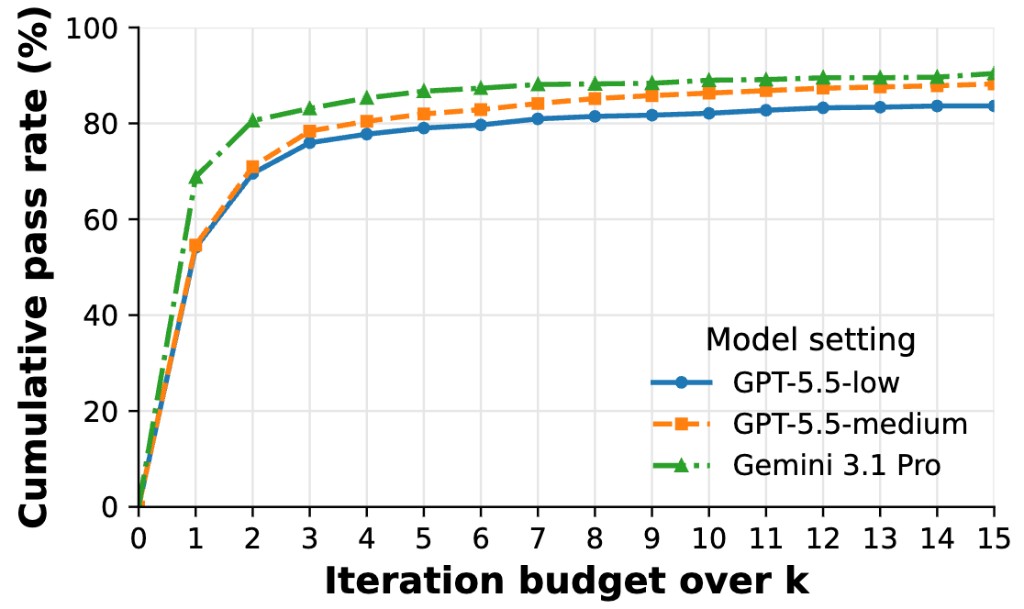}
\caption{Cumulative AxDafny pass rate on DafnyBench vs. iteration budget across model settings. The primary feedback gains are heavily concentrated within the first 5 iterations.}
  \label{fig:dafnybench-pass-rate}
\end{figure}

Figure~\ref{fig:dafnybench-pass-rate} shows that verifier-feedback gains on DafnyBench are concentrated in the early iterations. The Gemini-3.1-Pro setting achieves the highest verification success, reaching 92.7\% (725/782), compared to the 88.9\% (695/782) of GPT-5.5 medium and 85.3\% of GPT-5.5 low. At a fixed budget of 15 iterations,
GPT-5.5 medium exceeds GPT-5.5 low by 3.6 percentage points, while Gemini Pro
exceeds GPT-5.5 low by 7.4 percentage points; by iteration 5, Gemini Pro
already surpasses the final GPT-5.5 low pass rate. 
The models demonstrated gains between 25\% and 33\% over the course of the 15 iterations.

\begin{figure}[t]
    \centering
    \includegraphics[width=\columnwidth]{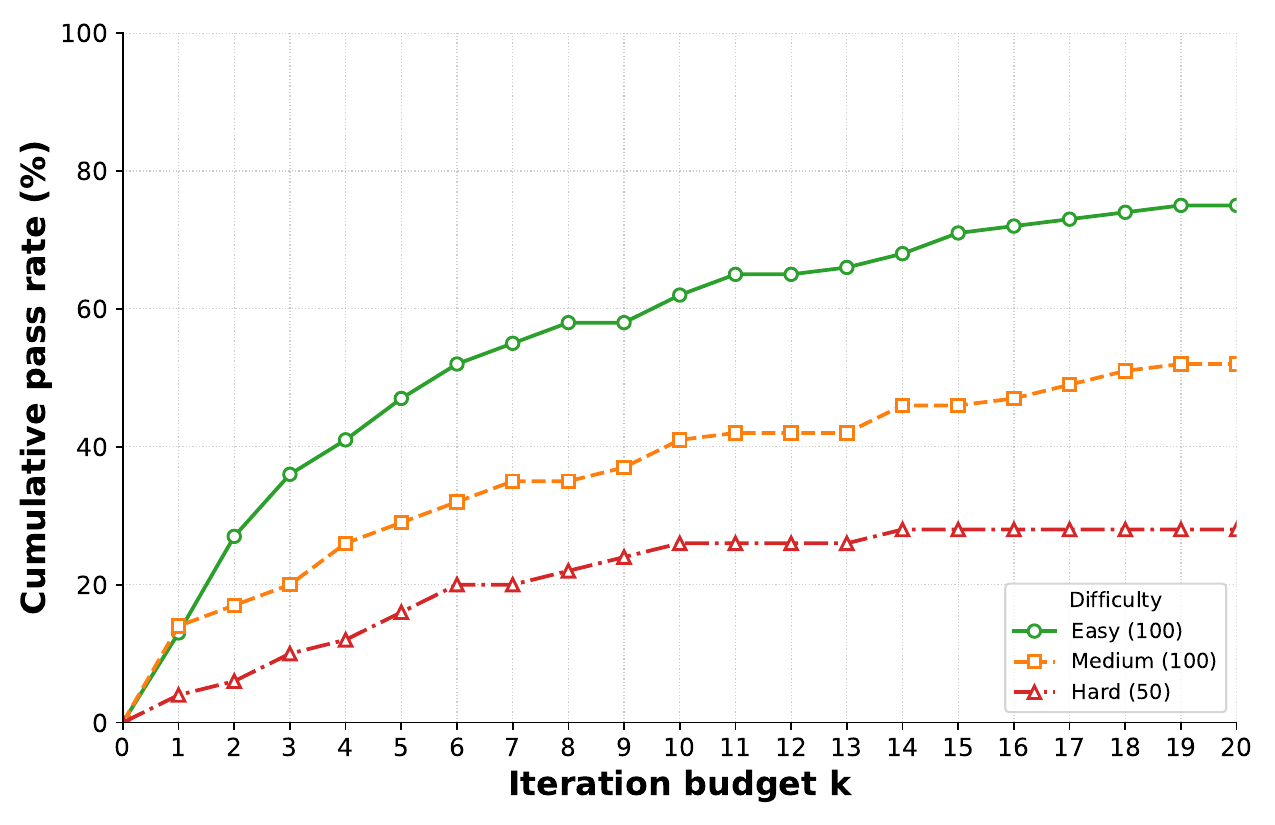}
    \caption{Cumulative AxDafny pass rate on LCB-Pro-Dafny across difficulty splits. Accuracy scales steadily with an increased iteration budget across all levels. Compared to proof-hint filling on DafnyBench, end-to-end program synthesis derives more sustained benefits from more iterations.}
    \label{fig:lcbpro-dafny-pass-rate}
\end{figure}

Table~\ref{tab:lcb-pro-dafny-results} and
Figure~\ref{fig:lcbpro-dafny-pass-rate} show that AxDafny substantially
improves over the GPT-5.5 pass@1 baseline on LCB-Pro-Dafny. However, the hard
split remains far from saturated: AxDafny verifies only 28.0\% of hard
instances. This confirms that LCB-Pro-Dafny is a challenging benchmark for
verified program synthesis.

\begin{table}[!t]
    \caption{Results on LCB-Pro-Dafny by difficulty split. Each entry is the
    verification success rate (\%).}
    \label{tab:lcb-pro-dafny-results}
    \centering
    \begin{tabular}{lcccc}
      \toprule
      Method & Easy & Medium & Hard & Overall \\[-2pt]
             & {\scriptsize $n{=}100$} & {\scriptsize $n{=}100$} & {\scriptsize $n{=}50$} & {\scriptsize $n{=}250$} \\
      \midrule
      \textbf{AxDafny}  & \textbf{75.0\%} & \textbf{52.0\%} & \textbf{28.0\%} & \textbf{56.4\%} \\
      GPT-5.5 pass@1 & 13.0\% & 14.0\% &  4.0\% & 11.6\% \\
      \bottomrule
    \end{tabular}
  \end{table}

\paragraph{Executable correctness.}
We also compile verified Dafny outputs to Python and evaluate them with the
original LCB-Pro test harness. This secondary evaluation is only defined for
verified solutions: unverified Dafny files do not build. On the easy split,
AxDafny verifies 75/100 tasks; among these verified outputs, 32 pass the
original executable tests, 39 fail by time limit exceeded (TLE), and 4 fail by
memory limit exceeded (MLE). On the medium split, AxDafny verifies 52/100 tasks;
among these verified outputs, 6 pass the executable tests, 44 fail by TLE, and 2
fail by MLE. For context, Appendix~\ref{app:lcb-pro-standard-results}
reports standard LCB-Pro leaderboard results under the original executable-code
evaluation.

\section{Discussion}

The verification results show two complementary outcomes. On DafnyBench,
AxDafny verifies 92.7\%, improving proof-hint
synthesis over published baselines and indicating that verifier-guided repair
is effective for generating proof annotations when the implementation and specification are fixed. On LCB-Pro-Dafny, AxDafny improves
substantially over direct GPT-5.5 pass@1 generation, but the hard split remains
far from saturated. This suggests that LCB-Pro-Dafny is a challenging testbed for end-to-end
verified program synthesis and a useful benchmark for future work on
verifier-guided coding agents.

The secondary executable evaluation measures a different property: whether
verified Dafny programs also satisfy the runtime and input-output constraints
of the original LCB-Pro test harness. Among verified outputs that fail this
secondary evaluation, the observed failures are resource-limit failures (TLE or
MLE), rather than wrong-answer failures or failures to compile unverified code.
These failures have two
main sources. First, the Dafny specifications
constrain functional correctness but generally do not constrain asymptotic
complexity or memory usage. Agents can therefore produce implementations
that are easier to verify but inefficient at runtime. Second, Dafny
compilation to Python relies on runtime implementations for language
constructs such as sets and maps, which can be slower than native 
Python or C++ implementations. We partially mitigate this second source
with a lightweight Python compatibility layer for selected Dafny runtime constructs.

These results show that verifier acceptance and competitive-programming
performance remain distinct objectives. A second limitation is the relative
scarcity of Dafny training data compared with mainstream programming
languages. As a rough proxy for this gap, GitHub code search returns
\(4 \times 10^8\) Python files and \(1.2 \times 10^5\) Dafny
files,\footnote{GitHub code search:
\url{https://github.com/search?type=code&q=language:Python} and
\url{https://github.com/search?type=code&q=language:Dafny}, accessed
2026-06-26.} a difference of more than three orders of magnitude.

\medskip

\section*{Software and Data}

We release the code, benchmark data, and evaluation harness in a public
GitHub repository: \url{https://github.com/Axiomatic-AI/ax-dafny}. For LCB-Pro-Dafny, we release the problem statements, Dafny specifications, and evaluation
harness. The harness checks Dafny verification, compiles verified Dafny
solutions to Python, and evaluates the resulting programs against the
LiveCodeBench-Pro test harness \citep{zheng2026livecodebenchpro} under
fixed time limits. We also provide a lightweight Python compatibility layer for selected Dafny runtime constructs to reduce Time Limit Exceeded (TLE) failures during evaluation.

\section*{Acknowledgments}

We thank Krystian Nowakowski and Matteo Cipollina for helpful discussions and
feedback during the development of this work.

\section*{Impact Statement}

Formal verification can make everyday coding tasks more reliable by
checking program behavior against explicit specifications rather than only
against finite test suites. If tools for verified code generation become
more usable, developers could apply stronger correctness checks to routine
library functions, data-processing code, and infrastructure components
where subtle edge cases are difficult to test exhaustively.

At the same time, verified code generation can create misplaced confidence
when specifications are incomplete, incorrect, or misaligned with user
intent. Our evaluation therefore treats the specification as a trusted
object and explicitly checks for specification modification and vacuous
proof shortcuts. These safeguards are necessary for using formal
verification in practical coding workflows.


\bibliographystyle{icml2026}
\bibliography{references}

@article{loughridge2024dafnybench,
  title = {DafnyBench: A Benchmark for Formal Software Verification},
  author = {Loughridge, Chloe and Sun, Qinyi and Ahrenbach, Seth and
            Cassano, Federico and Sun, Chuyue and Sheng, Ying and
            Mudide, Anish and Misu, Md Rakib Hossain and Amin, Nada and
            Tegmark, Max},
  journal = {arXiv preprint arXiv:2406.08467},
  year = {2024}
}

@article{misu2024towards,
  title = {Towards AI-Assisted Synthesis of Verified Dafny Methods},
  author = {Misu, Md Rakib Hossain and Lopes, Cristina V. and Ma, Iris and
            Noble, James},
  journal = {Proc. ACM Softw. Eng.},
  volume = {1},
  number = {FSE},
  year = {2024},
  doi = {10.1145/3643763},
  url = {https://doi.org/10.1145/3643763}
}

@inproceedings{leino2010dafny,
  title = {Dafny: An Automatic Program Verifier for Functional Correctness},
  author = {Leino, K. Rustan M.},
  booktitle = {Logic for Programming, Artificial Intelligence, and Reasoning},
  pages = {348--370},
  publisher = {Springer},
  year = {2010}
}

@article{requena2026minimalagentautomatedtheorem,
  title = {A Minimal Agent for Automated Theorem Proving},
  author = {Requena Pozo, Borja and Letson, Austin and Nowakowski, Krystian and
            Beltran Ferreiro, Izan and Sarra, Leopoldo},
  journal = {arXiv preprint arXiv:2602.24273},
  year = {2026},
  eprint = {2602.24273},
  archivePrefix = {arXiv}
}

@article{yang2024sweagent,
  title = {{SWE-agent}: Agent-Computer Interfaces Enable Automated Software
           Engineering},
  author = {Yang, John and Jimenez, Carlos E. and Wettig, Alexander and
            Lieret, Kilian and Yao, Shunyu and Narasimhan, Karthik and
            Press, Ofir},
  journal = {arXiv preprint arXiv:2405.15793},
  year = {2024},
  eprint = {2405.15793},
  archivePrefix = {arXiv},
  primaryClass = {cs.SE}
}

@misc{banerjee2026dafnypro,
  title = {DafnyPro: LLM-Assisted Automated Verification for Dafny Programs},
  author = {Banerjee, Debangshu and Bouissou, Olivier and Zetzsche, Stefan},
  year = {2026},
  eprint = {2601.05385},
  archivePrefix = {arXiv},
  primaryClass = {cs.SE}
}

@misc{poesia2024dafnyannotator,
  title = {dafny-annotator: AI-Assisted Verification of Dafny Programs},
  author = {Poesia, Gabriel and Loughridge, Chloe and Amin, Nada},
  year = {2024},
  eprint = {2411.15143},
  archivePrefix = {arXiv},
  primaryClass = {cs.SE}
}

@article{xu2025dafnycomp,
  title = {Local Success Does Not Compose: Benchmarking Large Language Models
           for Compositional Formal Verification},
  author = {Xu, Xu and Li, Xin and Qu, Xingwei and Fu, Jie and Yuan, Binhang},
  journal = {arXiv preprint arXiv:2509.23061},
  year = {2025}
}

@inproceedings{zeng2026veriequivbench,
  title = {VeriEquivBench: An Equivalence Score for Ground-Truth-Free
           Evaluation of Formally Verifiable Code},
  author = {Zeng, Lingfei and Che, Fengdi and Huang, Xuhan and Ye, Fei and
            Xu, Xu and Yuan, Binhang and Fu, Jie},
  booktitle = {International Conference on Learning Representations},
  year = {2026},
  eprint = {2510.06296},
  archivePrefix = {arXiv},
  primaryClass = {cs.PL}
}

@misc{jetbrains2024humanevaldafny,
  title = {HumanEval-Dafny: Translating HumanEval to Dafny},
  author = {{JetBrains Research}},
  year = {2024},
  howpublished = {\url{https://github.com/JetBrains-Research/HumanEval-Dafny}}
}

@article{chen2021evaluating,
  title = {Evaluating Large Language Models Trained on Code},
  author = {Chen, Mark and Tworek, Jerry and Jun, Heewoo and Yuan, Qiming and
            Pinto, Henrique Ponde de Oliveira and Kaplan, Jared and Edwards,
            Harri and Burda, Yuri and Joseph, Nicholas and Brockman, Greg and
            Ray, Alex and Puri, Raul and Krueger, Gretchen and Petrov, Michael
            and Khlaaf, Heidy and Sastry, Girish and Mishkin, Pamela and Chan,
            Brooke and Gray, Scott and Ryder, Nick and Pavlov, Mikhail and
            Power, Alethea and Kaiser, Lukasz and Bavarian, Mohammad and
            Winter, Clemens and Tillet, Philippe and Such, Felipe Petroski and
            Cummings, Dave and Plappert, Matthias and Chantzis, Fotios and
            Barnes, Elizabeth and Herbert-Voss, Ariel and Guss, William Hebgen
            and Nichol, Alex and Paino, Alex and Tezak, Nikolas and Tang, Jie
            and Babuschkin, Igor and Balaji, Suchir and Jain, Shantanu and
            Saunders, William and Hesse, Christopher and Carr, Andrew N. and
            Leike, Jan and Achiam, Josh and Misra, Vedant and Morikawa, Evan
            and Radford, Alec and Knight, Matthew and Brundage, Miles and
            Murati, Mira and Mayer, Katie and Welinder, Peter and McGrew, Bob
            and Amodei, Dario and McCandlish, Sam and Sutskever, Ilya and
            Zaremba, Wojciech},
  journal = {arXiv preprint arXiv:2107.03374},
  year = {2021}
}

@article{austin2021program,
  title = {Program Synthesis with Large Language Models},
  author = {Austin, Jacob and Odena, Augustus and Nye, Maxwell and Bosma,
            Maarten and Michalewski, Henryk and Dohan, David and Jiang, Ellen
            and Cai, Carrie and Terry, Michael and Le, Quoc and Sutton, Charles},
  journal = {arXiv preprint arXiv:2108.07732},
  year = {2021}
}

@article{zheng2025livecodebenchpro,
  title = {LiveCodeBench Pro: How Do Olympiad Medalists Judge LLMs in
           Competitive Programming?},
  author = {Zheng, Zihan and Cheng, Zerui and Shen, Zeyu and Zhou, Shang and
            Liu, Kaiyuan and He, Hansen and Li, Dongruixuan and Wei, Stanley and
            Hao, Hangyi and Yao, Jianzhu and Sheng, Peiyao and Wang, Zixuan and
            Chai, Wenhao and Korolova, Aleksandra and Henderson, Peter and
            Arora, Sanjeev and Viswanath, Pramod and Shang, Jingbo and
            Xie, Saining},
  journal = {arXiv preprint arXiv:2506.11928},
  year = {2025}
}

@misc{zheng2026livecodebenchpro,
  title = {LiveCodeBench-Pro: LLM Benchmarking Toolkit},
  author = {Zheng, Gavin},
  year = {2026},
  howpublished = {\url{https://github.com/GavinZhengOI/LiveCodeBench-Pro}}
}

@misc{livecodebenchpro2026leaderboard,
  title = {LiveCodeBench Pro Live Leaderboard},
  author = {{LiveCodeBench Pro}},
  year = {2026},
  howpublished = {\url{https://livecodebenchpro.com/projects/livecodebench-pro/leaderboard}},
  note = {Accessed 2026-05-23}
}

\newpage
\appendix
\onecolumn
\section{LCB-Pro-Dafny Construction}
\label{app:lcb-pro-dafny-construction}

LCB-Pro-Dafny was constructed from LiveCodeBench-Pro competition problems
\citep{zheng2025livecodebenchpro}. We retained the original difficulty labels
and selected 250 instances: 100 easy, 100 medium, and 50 hard. Each instance was
converted into a Dafny task containing the natural-language problem statement, a
Dafny method signature, and a formal specification.

The initial Dafny specifications were produced with model assistance and then
manually reviewed. During review, we removed or revised tasks with inconsistent,
underspecified, or vacuous specifications, and checked that the remaining
specifications captured the intended functional behavior of the original
problem. The evaluation harness treats these specifications as fixed: submitted
solutions may add helper definitions and proof artifacts, but may not weaken or
rewrite the provided method signature, definitions, or specifications.

The harness runs Dafny verification as the primary acceptance criterion and
applies additional checks for proof-bypass constructs such as \texttt{assume},
\texttt{\{:extern\}}, and \texttt{\{:verify false\}}. For the secondary
executable evaluation, verified Dafny solutions are compiled to Python and run
against the corresponding LiveCodeBench-Pro tests under the original time
limits.

\paragraph{Validation checks.}
We report the validation outcomes used during benchmark curation. All 250
released specifications parse and type-resolve with \texttt{dafny resolve
--allow-warnings} under Dafny 4.11.0. Static checks found no missing
\texttt{ensures} clauses and no trivially true postconditions. Under Dafny's
stricter default warning policy, 49 specifications emitted quantifier-trigger
warnings, but otherwise parsed and type-resolved without errors.

LLM-as-judge review against the original problem statements flagged 14/250
specifications (5.6\%) as semantic mismatches requiring fixes. In one
representative medium instance, an earlier specification draft represented a
set of chosen objects as a sequence but did not forbid duplicate entries; we
fixed this by requiring the sequence to be distinct. All flagged specifications
were revised and reviewed before inclusion.

As an additional vacuity check, no placeholder implementation verified, and no
automatically generated constant-or-empty-output mutation verified in the 234
cases where the mutation harness applied.

\section{Standard LCB-Pro Results}
\label{app:lcb-pro-standard-results}

Table~\ref{tab:lcb-pro-standard-results} reports standard LCB-Pro leaderboard
results under the benchmark's original executable-code evaluation. These
numbers provide context for the difficulty of the underlying programming
tasks, but they are not directly comparable to LCB-Pro-Dafny verification
rates: standard LCB-Pro measures whether generated Python code passes runtime
tests, while LCB-Pro-Dafny measures whether generated Dafny code verifies
against formal specifications. The comparison helps interpret our secondary
executable evaluation: strong coding models still struggle on the hard split
under ordinary runtime testing, while LCB-Pro-Dafny shifts the evaluation
target to producing machine-checked implementations.

\begin{table}[!t]
  \caption{Published LCB-Pro live leaderboard results from the project
  website. Easy, medium, and hard report current live success rates; overall
  is computed by aggregating reported pass counts across the live difficulty
  splits rather than using the website's Elo rating
  \citep{livecodebenchpro2026leaderboard}.}
  \label{tab:lcb-pro-standard-results}
  \centering
  \small
  \setlength{\tabcolsep}{2pt}
  \begin{tabular}{lcccc}
    \toprule
    Method & Easy & Medium & Hard & Overall \\
    \midrule
    GPT-5.2-high & 90.1 & 52.1 & 15.9 & 53.1 \\
    GPT-5-high & 90.1 & 40.8 & 4.2 & 45.1 \\
    o4-mini-high & 85.7 & 29.2 & 1.4 & 37.4 \\
    Qwen3 Next 80B A3B Thinking & 69.0 & 7.0 & 0.0 & 25.4 \\
    Claude 4.5 Sonnet Thinking & 53.5 & 0.0 & 0.0 & 17.8 \\
    DeepSeek R1 & 50.7 & 1.4 & 0.0 & 17.4 \\
    \bottomrule
  \end{tabular}
\end{table}

\section{Model Ablation}
\label{app:model-ablation}

\begin{figure}[!htbp]
  \centering
  \includegraphics[width=0.72\columnwidth]{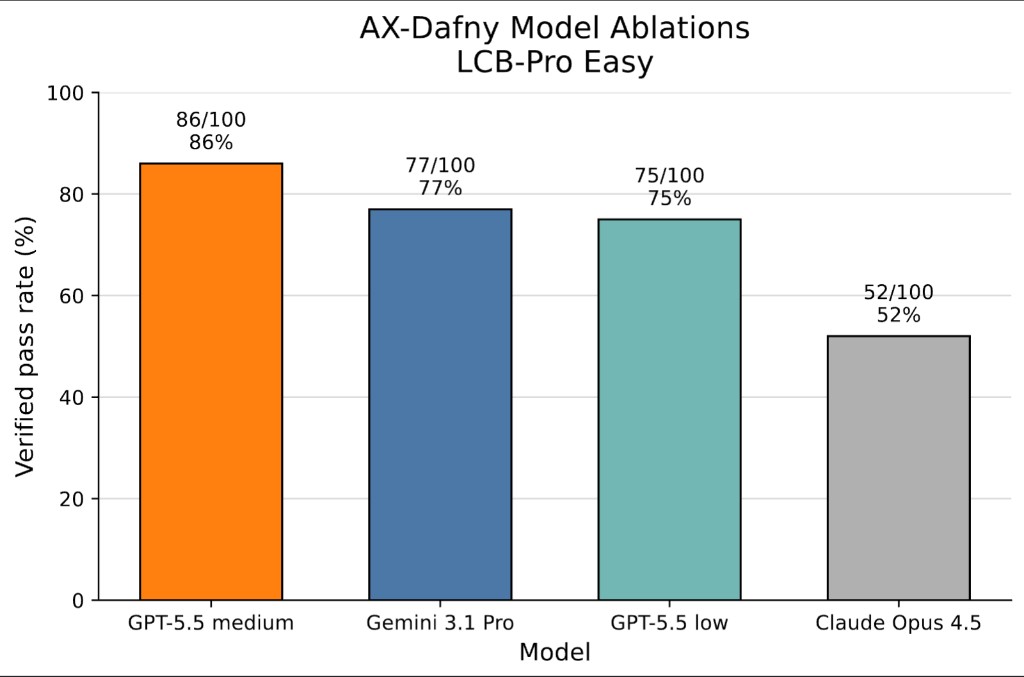}
  \caption{Model ablation on the LCB-Pro-Dafny easy split. Bars report verified
  pass rate over 100 easy instances.}
  \label{fig:lcb-pro-easy-model-ablation}
\end{figure}

Figure~\ref{fig:lcb-pro-easy-model-ablation} reports a small model ablation on
the LCB-Pro-Dafny easy split. The results compare verified pass rates for the
same verifier-guided synthesis setting across four frontier model settings.

The ablation suggests that performance on LCB-Pro-Dafny is sensitive to the
choice of base model and reasoning setting. On the easy split, GPT-5.5 medium
verifies 86/100 instances, Gemini 3.1 Pro verifies 77/100, GPT-5.5 low verifies
75/100, and Claude Opus 4.5 verifies 52/100.
We include this result to document model sensitivity for the verifier-guided
synthesis loop; broader model comparisons require evaluation across all
difficulty splits.


\end{document}